\documentclass[10pt,twocolumn,letterpaper]{article}

\usepackage[pagenumbers]{cvpr} 

\makeatletter
\robustify\@latex@warning@no@line
\makeatother
\usepackage{authblk}
\usepackage{acronym}
\usepackage{adjustbox}
\usepackage[accsupp]{axessibility} 
\usepackage{multirow}
\usepackage{siunitx}
\usepackage{nth}

\usepackage{array}
\newcolumntype{L}[1]{>{\raggedright\arraybackslash}m{#1}}
\newcolumntype{C}[1]{>{\centering\arraybackslash}p{#1}}

\usepackage{todonotes}
\usepackage{pifont}
\definecolor{cvprblue}{rgb}{0.21,0.49,0.74}
\usepackage[pagebackref,breaklinks,colorlinks,allcolors=cvprblue]{hyperref}
\usepackage{graphbox,graphicx}

\def\paperID{*****} 
\def\confName{CVPR}
\def\confYear{2025}

\title{RADLER: Radar Object Detection Leveraging Semantic 3D City Models and Self-Supervised Radar-Image Learning}

\author{Yuan Luo\textsuperscript{1,3}, Rudolf Hoffmann\textsuperscript{3}, Yan Xia\textsuperscript{*1,2}, Olaf Wysocki\textsuperscript{1}, Benedikt Schwab\textsuperscript{1}, Thomas H. Kolbe\textsuperscript{1}, Daniel Cremers\textsuperscript{1,2} \\
\textsuperscript{1}Technical University of Munich, \textsuperscript{2}MCML, \textsuperscript{3}GPP Communication GmbH\\
{\tt\small yan.xia@tum.de\textsuperscript{*} corresponding author}}

\begin{document}
\maketitle
\acrodef{ra}[RA]{Range-Azimuth}
\acrodef{ssl}[SSL]{self-supervised learning}
\acrodef{mmwave}[mmWave]{millimeter-wave}
\acrodef{crctum}[]{RadarCity}
\acrodef{citygml}[CityGML]{City Geography Markup Language}
\acrodef{ap}[AP]{average precision}
\acrodef{ar}[AR]{average recall}
\acrodef{map}[mAP]{mean average precision}
\acrodef{mar}[mAR]{mean average recall}
\acrodef{ra}[RA]{range-azimuth}
\acrodef{confmaps}[ConfMaps]{confidence maps}
\acrodef{lnms}[L-NMS]{location-based non-maximum suppression}
\acrodef{cfar}[CFAR]{constant false alarm rate}
\acrodef{sdm}[SDM]{semantic-depth maps}
\acrodef{lod}[LOD]{levels of detail}
\acrodef{rodnet}[RODNet]{radar object detection network}
\acrodef{vit}[ViT]{Vision Transformer}
\acrodef{ols}[OLS]{object location similarity}
\acrodef{ffts}[FFTs]{Fast Fourier Transforms}
\acrodef{rf}[RF]{radio frequency}
\acrodef{fmcw}[FMCW]{frequency modulated continuous wave}
\acrodef{etl}[ETL]{Extract Transform Load}
\acrodef{mlp}[MLP]{multi-layer perceptron}
\acrodef{clip}[CLIP]{Contrastive Language–Image Pre-training}
\acrodef{bce}[BCE]{binary cross-entropy}
\acrodef{gt}[GT]{ground truth}
\acrodef{fps}[FPS]{frames per second}
\acrodef{fov}[FoV]{field of view}
\acrodef{gnn}[GNN]{graph neural network}
\acrodef{bev}[BEV]{bird's-eye-view}
\acrodef{snr}[SNR]{signal-to-noise ratio}
\acrodef{tdm}[TDM]{time division multiplexing}
\acrodef{mimo}[MIMO]{multiple-input multiple-output}
\acrodef{tum}[TUM]{Technical University of Munich}
\begin{abstract}
Semantic 3D city models are worldwide easy-accessible, providing accurate, object-oriented, and semantic-rich 3D priors. To date, their potential to mitigate the noise impact on radar object detection remains under-explored. 
In this paper, we first introduce a unique dataset, RadarCity, comprising 54K synchronized radar-image pairs and semantic 3D city models.  
Moreover, we propose a novel neural network, RADLER, leveraging the effectiveness of contrastive \ac{ssl} and semantic 3D city models to enhance radar object detection of pedestrians, cyclists, and cars. 
Specifically, we first obtain the robust radar features via a \ac{ssl} network in the radar-image pretext task. We then use a simple yet effective feature fusion strategy to incorporate semantic-depth features from semantic 3D city models. Having prior 3D information as guidance, RADLER obtains more fine-grained details to enhance radar object detection. 
We extensively evaluate RADLER on the collected RadarCity dataset and demonstrate average improvements of 5.46\% in \ac{map} and 3.51\% in \ac{mar} over previous radar object detection methods.
We believe this work will foster further research on semantic-guided and map-supported radar object detection. 
Our project page is publicly available at \href{https://gpp-communication.github.io/RADLER}{https://gpp-communication.github.io/RADLER}.

\end{abstract}    
\section{Introduction}
\label{sec:intro}

\Ac{fmcw} radar is a widely used type of \ac{mmwave} radar, having applications in diverse domains, including object detection \cite{wang2021rodnet,nabati2021centerfusion,nabati2019rrpn,huang2024l4dr}, human activity detection \cite{singh2019radhar,li2021semisupervised}, and hand gesture recognition \cite{zhang2018latern,choi2019short}.
Unlike image and LiDAR object detection~\cite{xia2023lightweight, wu2024boosting}, \Ac{fmcw} radar produces data in \ac{rf} form. 
By applying \ac{ffts}, range, Doppler velocity, and azimuth can be derived, enabling the generation of radar point clouds through \ac{cfar} detection and clustering. 
However, radar signals are hardly interpretable by humans, making manual annotation impractical. The lack of labeled radar datasets significantly limits the development of learning-based methods in radar object detection.

\begin{figure}[t]
    \centering
    \includegraphics[width=\linewidth]{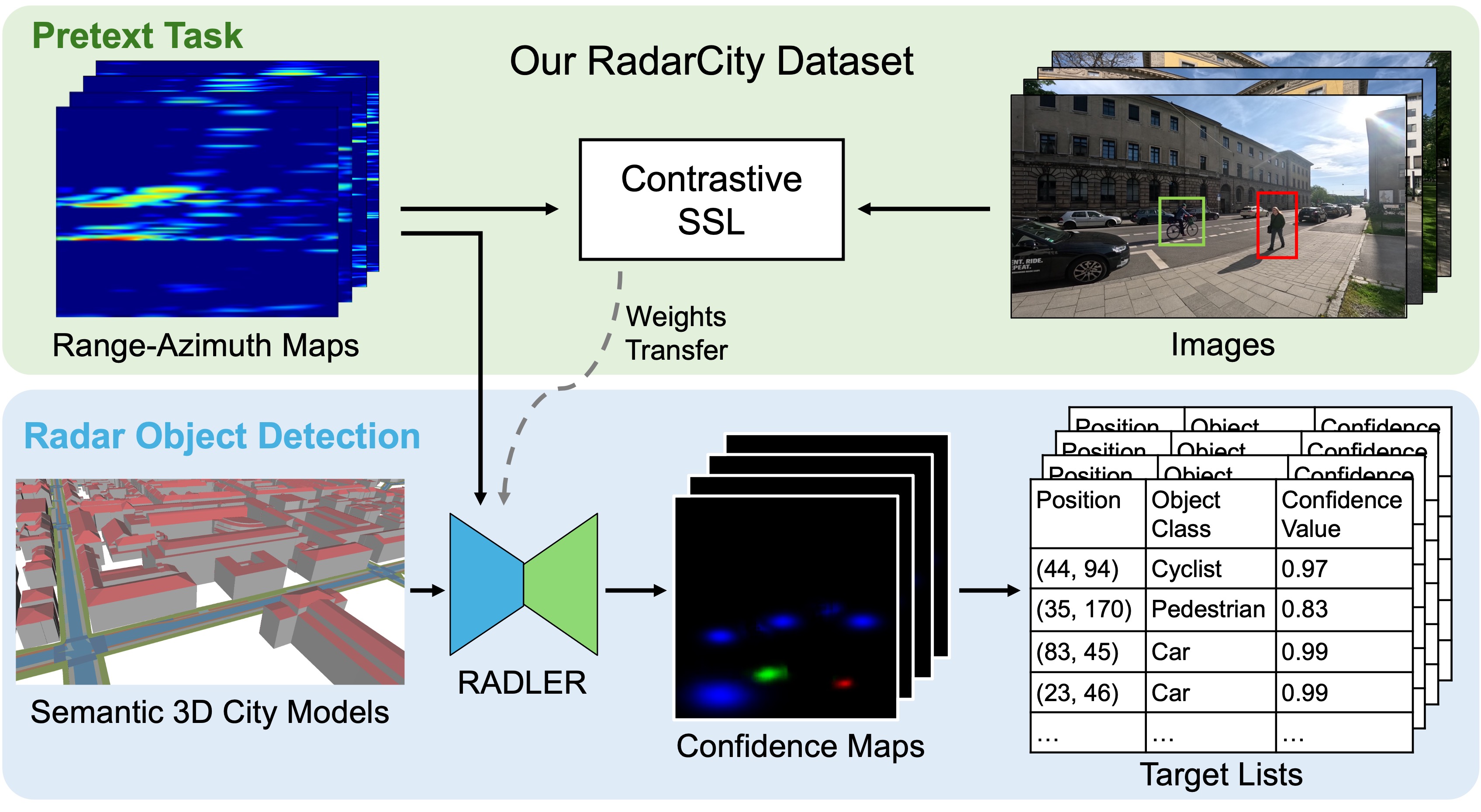}
    \caption{The workflow of RADLER. Representations of the \ac{ra} maps are learned against the images in the pretext task and transferred to the radar object detection task, enhanced by prior information from semantic 3D city models.}
    \label{fig:teaser}
\end{figure}

Recent contrastive \ac{ssl} has demonstrated its promise of learning directly from data without annotation in image-related tasks \cite{chen2020simple,he2020momentum,grill2020bootstrap}. 
The inherent spatial or temporal relationship in cross-modal data, such as radar-image pairs, makes them suitable for contrastive \ac{ssl} approaches. 
Recent studies \cite{hao2024bootstrapping,alloulah2022self,alloulah2023look} have supported its effectiveness in radar object detection. 
However, \ac{fmcw} radars are prone to environmental clutter and multipath propagation, resulting in ghost objects and masking effects that adversely affect accurate detection. 
While some methods~\cite{chamseddine2021ghost,liu2020multipath,kraus2020using} address these issues for radar point clouds, few tackle them in the \ac{rf} domain.

Besides, map-based radar object detection methods are scarce and typically rely only on 3D mesh-like geometries without incorporating semantic information \cite{hong2020radarslam}. 
Actually, semantic 3D city models have become increasingly widespread and easily accessible, e.g., more than 215 million building models worldwide available at no cost \cite{wysocki2024reviewing,awesomeCityGML}. 
Such city models can provide rich contextual information, such as building functions, road structures, and separated traffic lanes, which can enhance radar perception. 
Despite their availability, existing radar datasets do not integrate semantic 3D city models, limiting research into their potential benefits.  

To bridge this gap, we first introduce RadarCity, a new dataset comprising 54K synchronized radar-image pairs enriched with semantic 3D city models at Technical University of Munich, Germany. 
The data are filtered to contain only pedestrians, cyclists, and cars. 
Furthermore, we propose RADLER, a novel RADar object detector, LEveraging readily available semantic 3D city models as prior information and a \ac{ssl} strategy, as shown in \cref{fig:teaser}.

Inspired by \citet{alloulah2022self}, we first propose a pretext radar-image task to learn robust radar representations against the images using a contrastive \ac{ssl} strategy. Then, we apply a simple yet effective fusion strategy to insert the semantic and depth information from the easily accessible semantic 3D city models. Specifically, we process \ac{ra} maps and classify objects via \ac{confmaps}, with the optional fusion of semantic and depth information from \ac{sdm}. 
Experiments on our RadarCity dataset show that our approach achieves a \ac{map} of 90.69\% and a \ac{mar} of 95.28\% without \ac{sdm}, with performance being further improved by 4.17\% and 0.67\%, respectively, when \ac{sdm} are incorporated. RADLER demonstrates superior detection accuracy compared to baseline \ac{rodnet} models, highlighting the benefits of integrating semantic 3D city models with radar data.

Our contributions are summarized as follows:
\begin{itemize}

    \item To the best of our knowledge, we are the first to leverage easily accessible and standardized semantic 3D city models to assist radar-based object detection. To support this, we set up a new dataset, RadarCity, comprising 54K synchronized radar-image pairs with semantic 3D city models being used to generate \ac{sdm} for all scenes.
        
    \item We introduce a radar object detection method based on contrastive \ac{ssl}, RADLER, which effectively learns representations from radar data and improves detection accuracy.

    \item We enhance radar object detection performance by effectively incorporating semantic and geometric information from semantic 3D city models.

\end{itemize}
\section{Related Works}
\label{sec:related_work}

\noindent
\textbf{Contrastive self-supervised learning} 
Contrastive \ac{ssl} captures comprehensive feature representations by exploring not only a single modality but the inherent spatial and temporal relationships between different data modalities.
SimCLR represents a simple framework for contrastive \ac{ssl} \cite{chen2020simple}. It unleashes the potential of representation learning from large-scale image data without annotations. 
MoCo \cite{he2020momentum} introduced a dynamic dictionary incorporating a queue and momentum-updated parameters, significantly reducing hardware requirements while maintaining competitive performance in representation learning. Different downstream tasks in computer vision, such as object detection \cite{liu2020self, yang2021instance, baek2020psynet}, semantic segmentation \cite{hoyer2021three, ziegler2022self}, and 3D vision \cite{achituve2021self, xia2024text2loc, sanghi2020info3d}, also leverage the representations learned to achieve exceptional performance.
\Ac{clip} applies contrastive \ac{ssl} to text-image pairs to caption images \cite{radford2021learning}. In \cite{alwassel2020self}, videos are clustered based on their audio information using a contrastive \ac{ssl} model trained on video-audio data. Additionally, Text2Loc \cite{xia2024text2loc} establishes correspondences between 3D point clouds of objects and textual descriptions using a contrastive learning approach.

\noindent
\textbf{Contrastive self-supervised radar-image learning} 
Owing to the temporal relationship between radar and image data, utilizing contrastive \ac{ssl} is an ideal method to combine them. In \cite{alloulah2022self}, the authors apply contrastive \ac{ssl} on radar-image data from the CRUW dataset. The learned representations of different object classes on the \ac{ra} maps exhibit distinct clusters when visualized with t-SNE. The effectiveness of contrastive \ac{ssl} on radar-image data is further validated in \cite{alloulah2023look} using both synthetic and empirical datasets.  
However, the intrinsic differences still exist between image and radar data. Recent \cite{hao2024bootstrapping} implemented and evaluated radar-specific data augmentation methods to address this issue.

\noindent
\textbf{Interoperable semantic 3D city models}
To structure semantic, geometric, topological, and appearance information for cities and landscapes, the \ac{citygml} standard has become established internationally \cite{kolbeOGCCityGeography2021}.
The standard defines a conceptual data model and is issued by the Open Geospatial Consortium (OGC).
\Ac{citygml} is based on the standards from the ISO 191XX series of geographic information standards, which are widely used by geographic information systems.
Due to the absolute georeferencing with standardized spatial reference systems, accurate distance and area calculations can be carried out for entire cities and countries while taking the curvature of the earth into account.

In CityGML version 3.0, objects can be geometrically represented at \ac{lod} 0 to 3.
For example, building models in \ac{lod}0 are typically represented by their footprint, and in \ac{lod}1, by prismatic 3D solids. 
While \ac{lod}2 building models comprise realistic but generalized roofs and facades, \ac{lod}3 include detailed roof and facade structures.
At the granularity level \textit{lane}, the street space is decomposed into lane-specific traffic spaces and areas with predecessor-successor information \citep{kutznerCityGMLNewFunctions2020,beilApplicationsSemantic3D2024}.
The complete building stock in \ac{lod}2 of Germany, Poland, Switzerland, and large parts of Japan is provided and maintained with stable identifiers as open \ac{citygml} datasets by public authorities.
Since these datasets are commonly based on the official cadastre, the georeferencing accuracy is in the centimeter range \cite{RoschlaubBatscheider}.
As of 2024, a total of 216.5 million building models are available as open \ac{citygml} datasets worldwide, with an increasing number of transportation models \cite{wysocki2024reviewing}.
The current trends also show that more data is expected to be released owing to open data policies and novel reconstruction methods \cite{wysocki2023scan2lod3,xu2024roof,wang2024framework,lo2024roofdiffusion,tang2025texture2lod3}.

\section{Our RADLER}
\label{sec:method}

RADLER's workflow is illustrated in \cref{fig:model-architecture}. \Ac{ra} maps are
processed using the contrastive \ac{ssl} framework, MoCo \cite{he2020momentum}, in the pretext task to learn comprehensive representations against the images. These representations are transferred to the radar object detection task, where they can be fused with semantic and depth information, stored as \ac{sdm}, from semantic 3D city models. RADLER outputs \ac{confmaps} to classify and localize the objects. \Ac{lnms} is applied on the \ac{confmaps} to extract target lists, which are evaluated against \ac{gt} for \ac{ap} and \ac{ar}.

\subsection{Prior Information Selection}
To determine which prior information from the \ac{citygml} model could potentially enhance radar object detection, we analyze the movement pattern of objects on the \ac{ra} map. Consecutive \ac{ra} maps are animated to track the trajectories of the moving objects, represented as lines in distinct colors.

\begin{figure}[tbh]
    \centering
    \includegraphics[width=\linewidth]{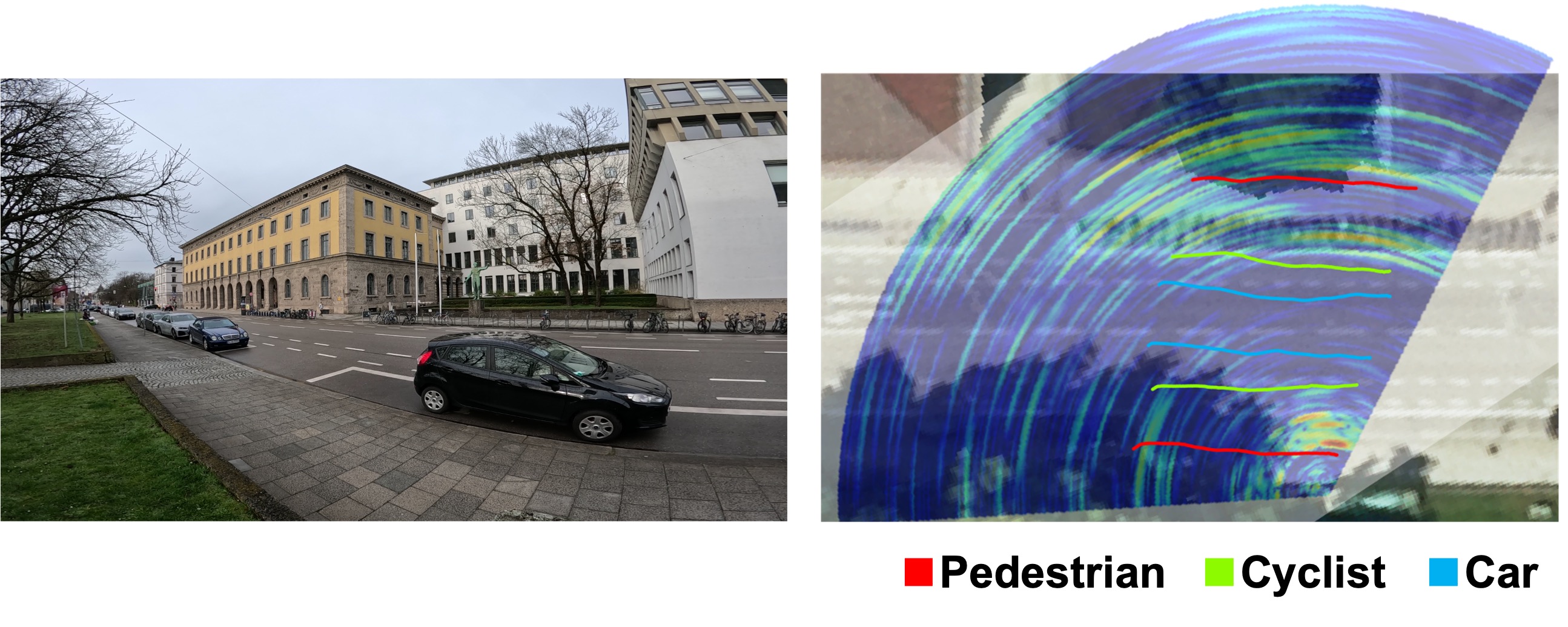}
    \caption{Tracking moving objects' trajectories on \ac{ra} maps: street-level camera view (left) and the \ac{bev} with trajectories on the \ac{ra} map (right).}
    \label{fig:track-line}
\end{figure}

The results, shown in \cref{fig:track-line}, demonstrate the accurate spatial registration of the tracked objects' trajectories by superimposing the \ac{ra} map onto a scaled and oriented \ac{bev} of the street. 
The \ac{ra} map's origin is precisely aligned with the sensor's deployment point with an offset of 10 centimeters, ensuring accurate georeferencing. 
Notably, the trajectories of different objects are observed to be nearly parallel, reflecting the linear nature of the traffic flow. 
Furthermore, these trajectories accurately align with the actual traffic lanes depicted in the \ac{bev}, confirming their correct spacing and distance from the sensor's deployment point. 

This precise alignment validates the reliability of the \ac{ra} map in representing the movement of objects within the street environment. 
We anticipate enhancing this representation by incorporating prior information about the street layout derived from semantic 3D city models, further improving the accuracy and contextual understanding of radar object detection.

Based on the process used to generate \cref{fig:track-line}, two types of information are identified as crucial for accurately constructing traffic lanes on the \ac{ra} map: \textbf{semantic and depth information}. Semantic information provides contextual details about different types of traffic objects, while depth information ensures traffic lanes' correct positioning and spacing from the device's perspective.

\subsection{Model Architecture and Design}
\begin{figure*}[tbh]
    \centering
    \includegraphics[width=\linewidth]{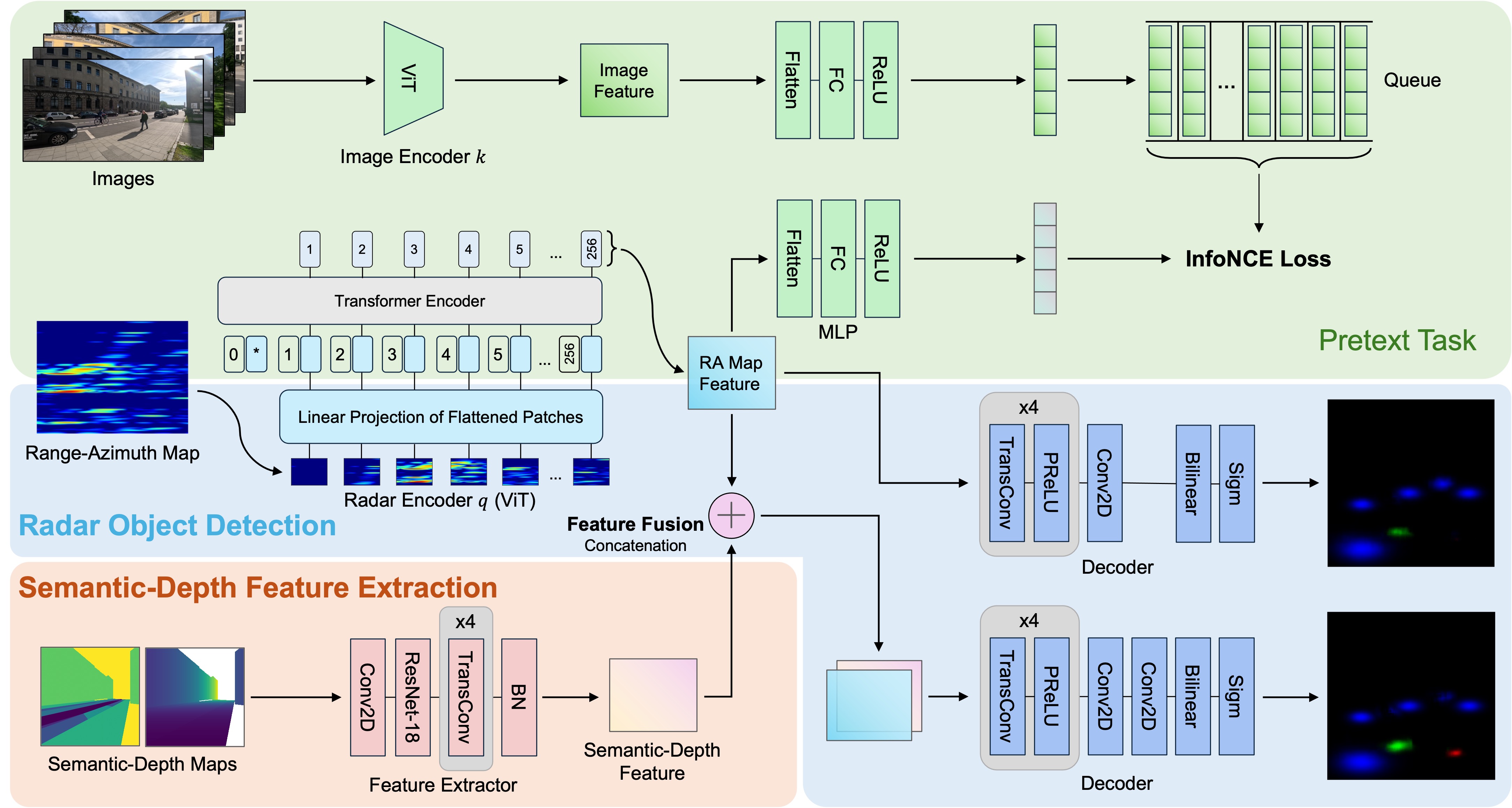}
    \caption{Schematic diagram of RADLER. The pretext task involves learning representations of \acs{ra} maps by contrasting them against corresponding image features to ensure semantic alignment \textit{(green)}. In the downstream radar object detection task, these learned representations are utilized by a decoder to generate \acs{confmaps} for object detection \textit{(blue)}. Additionally, the learned radar representations can be fused with features extracted from \acs{sdm} to further enhance detection performance \textit{(light orange)}.}
    \label{fig:model-architecture}
\end{figure*}

Contrastive \ac{ssl} involves two key stages: pretext and downstream tasks. The pretext task aims to establish comprehensive representations of the input data, which are then utilized in the downstream task for specific purposes. 

As shown in \cref{fig:model-architecture}, for the pretext task, we adopt MoCo \cite{he2020momentum} by using \ac{vit} \cite{dosovitskiy2020image} as the base encoder for both image and radar data. \ac{vit} is chosen due to its proven effectiveness in image-related tasks \cite{bao2023all,li2022exploring}, aligning well with the image-like nature of \ac{ra} maps. The encoders are initialized with weights pre-trained on ImageNet-1K data and fine-tuned during the training. The encoded features are further processed by a \ac{mlp} to reduce dimensionality before calculating the contrastive loss using the InfoNCE loss function:

\begin{equation}
    \mathcal{L}_{\text{InfoNCE}} = -\log \frac{\exp\left(\text{sim}(\mathbf{x}, \mathbf{z}_P)\right)}{\sum_{j=0}^{N} \exp\left(\text{sim}(\mathbf{x}, \mathbf{z}_j)\right)}
\end{equation}
where \(\mathbf{Z} = \{\mathbf{z}_j \mid j = 0, \dots, N\}\) denotes the image samples stored in the queue, with \(\mathbf{z}_P \in \mathbf{Z}\) representing the positive sample, and \(\mathbf{Z} \setminus \mathbf{z}_P\) are the negative samples. \(\mathbf{x}\) refers to the \ac{ra} maps.

The cosine similarity function, \(\text{sim}(\mathbf{x}, \mathbf{y}) := \mathbf{x}^\top\mathbf{y} / \tau \), is used to calculate the similarities between \ac{ra} maps and images, with \(\tau\) as the temperature hyperparameter. The InfoNCE loss function encourages our model to associate the representation of a \ac{ra} map more closely with its positive than negative images. 

During training, backpropagation updates the radar branch, while the parameters of the image branch are updated using the momentum mechanism:
\begin{equation}
    \theta_k(n) \leftarrow m\theta_k(n-1) + (1 - m)\theta_q(n)
\end{equation} Here, \(\theta_k\) represents the parameters of the encoder and \ac{mlp} on the image branch, and \(\theta_q\) denotes the parameters of the encoder and \ac{mlp} on the radar branch. \(m \in [0, 1)\) is the momentum coefficient that controls the smoothness of \(\theta_k\)'s updates. At step $n$, the parameters \(\theta_k\) are updated based on the previous parameters \(\theta_k (n - 1)\) and the current parameters \(\theta_q (n)\).

In the downstream task of radar object detection, the radar encoder trained during the pretext task is transferred. The encoded feature is decoded into \ac{confmaps}, with an option to incorporate semantic-depth features extracted from \ac{sdm} using ResNet-18 \cite{he2016deep} as the feature extractor. Each channel in the \ac{confmaps} corresponds to one object class, and each pixel value represents the likelihood of an object of that class being present at that location. The predicted \ac{confmaps} are compared with the \ac{gt} \ac{confmaps} using the \ac{bce} loss function:
\begin{equation}\label{math:bce-loss}
    \ell = - \sum_{\text{cls}} \sum_{i,j} D_{i,j}^{\text{cls}} \log \hat{D}_{i,j}^{\text{cls}} + \left(1 - D_{i,j}^{\text{cls}}\right) \log \left(1 - \hat{D}_{i,j}^{\text{cls}}\right)
\end{equation} here, \(D\) represents the \ac{gt} \ac{confmaps}. \(\hat{D}\) represents the predicted \ac{confmaps}. \(\text{cls}\) stands for the object's class label. \((i, j)\) is the pixel's coordinates.
\section{RadarCity Dataset}
\label{sec:dataset}

\begin{figure}[tbh]
    \centering
    \includegraphics[width=.9\linewidth]{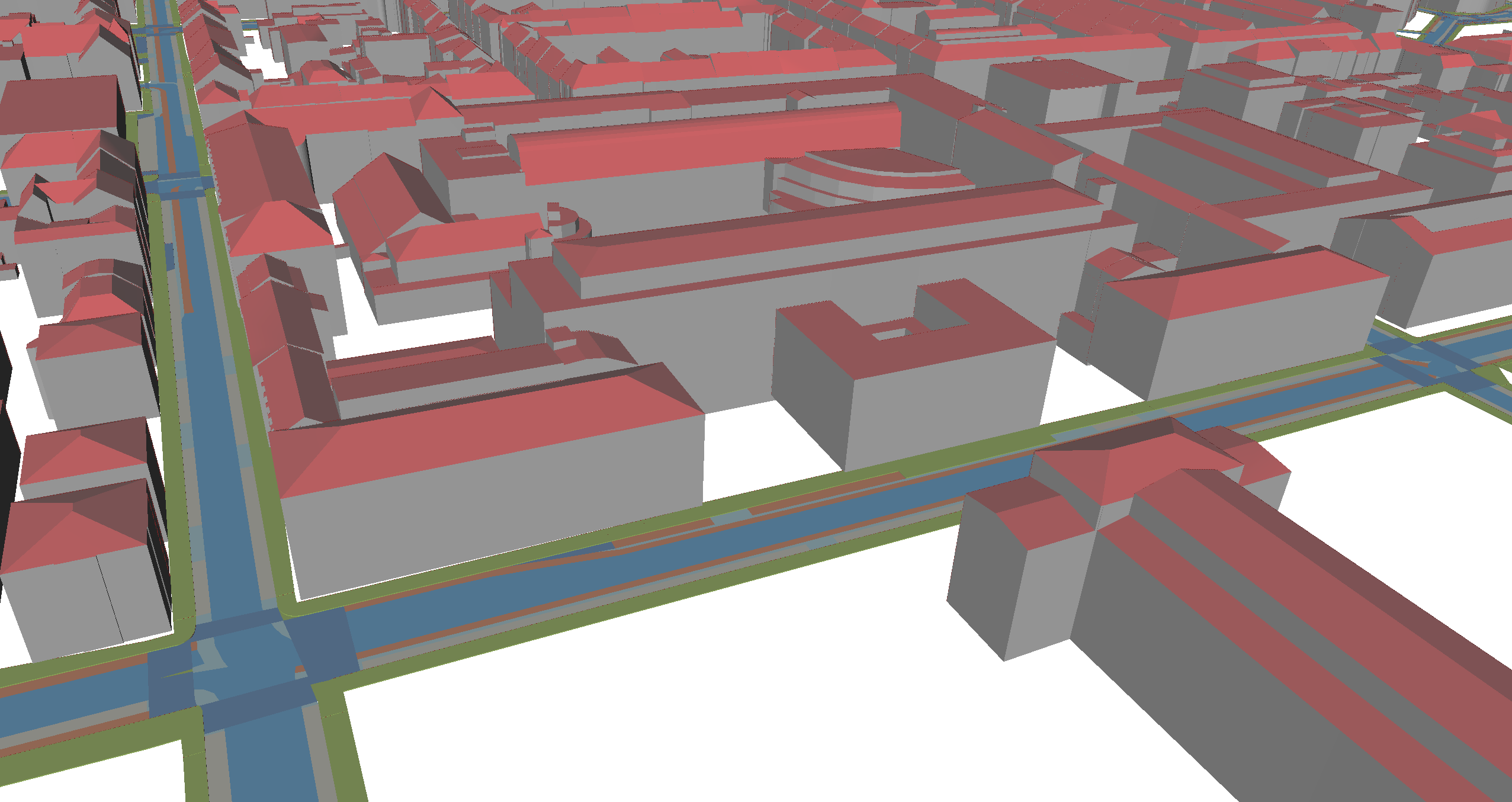}
    \caption{The utilized \acs{citygml} model representing \acs{tum}'s city center campus in Munich. 
    }
    \label{fig:tum-citygml}
\end{figure}

\begin{figure}[tbh]
    \centering
    \begin{subfigure}[t]{0.30875\linewidth}
        \centering
        \includegraphics[width=\linewidth]{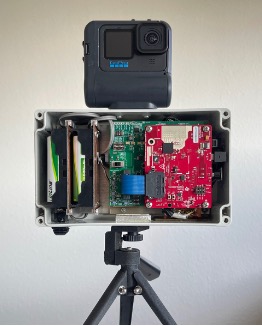}
        \caption{Sensor platform}
        \label{fig:device}
    \end{subfigure}
    \hfill
    \begin{subfigure}[t]{0.68125\linewidth}
        \centering
        \includegraphics[width=\linewidth]{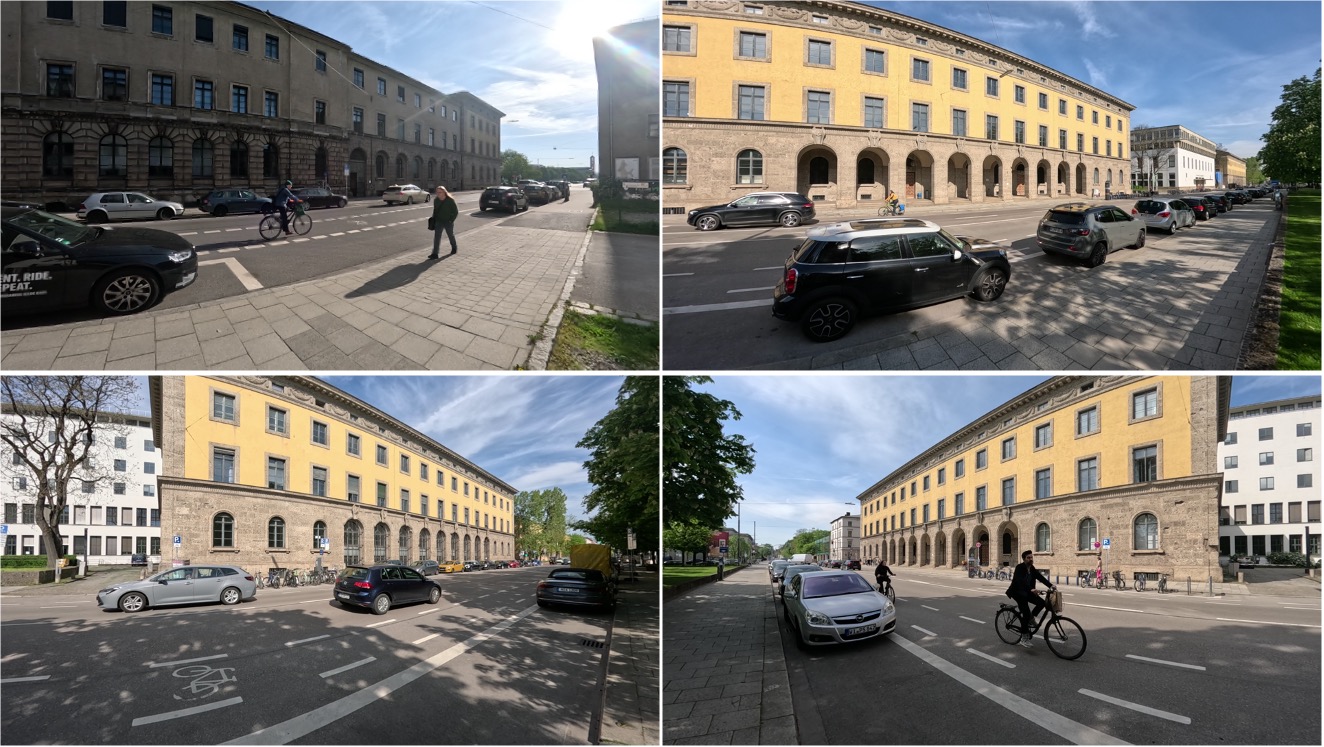}
        \caption{Camera images of four scenes}
        \label{fig:sample-data}
    \end{subfigure}
    \caption{Demonstration of the sensor platform used and all scenes for data collection in the RadarCity dataset.}
\end{figure}

The lack of support for semantic 3D city models in existing radar-image datasets necessitates the creation of a dedicated dataset for our research. 
For the area around the \ac{tum} university campus, comprehensive \ac{citygml} models exist, as shown in \cref{fig:tum-citygml}: While \ac{lod}2 building models are provided by the state mapping agency for the entire country, lane-level street models, tree models, and \ac{lod}3 building models are available within the TUM2TWIN project \cite{tum2twin}.
Consequently, we collect the RadarCity dataset there using the setup shown in \cref{fig:device}, 
which includes a camera and a \SI{77}{\giga\Hz} \ac{fmcw} radar. 
The AWR1843Boost radar has been widely used in research due to its high integration, affordability, strong performance, and comprehensive development support. 
To suit the requirements of our use case, we adapt the waveform parameters to achieve an unambiguous range of \SI{37.38}{\meter}. 
Moreover, we use an effective chirp bandwidth of \SI{1024.6}{\mega\Hz}, with $255$ chirps per frame and \ac{tdm} \ac{mimo} coding on the two transmit and four receive antennas to obtain a virtual linear horizontal array of eight receive antennas. 
This leads to a range resolution of \SI{14.6}{\centi\meter} and an angular resolution of \SI{15}{\degree}. 
The optimal standard deviations for a target with \SI{10}{\dB} \ac{snr} are \SI{3.6}{\centi\meter} for the range and \SI{2}{\degree} for the angle estimation.

The data collection is conducted stationarily, resulting in 54K camera and radar data pairs across four scenes. The data are manually filtered to include only pedestrians, cyclists, and cars. Sample data are visualized in \cref{fig:sample-data}. Contrastive \ac{ssl} alleviates the need for annotating data for the pretext task, which takes up 80\% of the data. The remaining 20\% is annotated manually for the downstream radar object detection task. This annotated portion is further split into training and testing datasets using an 80\%-20\% ratio.

\section{Experiments}
\label{sec:experiments}

\subsection{Prior Information Extraction}

To extract semantic and depth information, the objects in the \ac{citygml} models are grouped based on their semantic attributes. For instance, traffic lanes are categorized by lane types, with lanes of the same type grouped and stored individually.

\begin{figure}[tbh]
    \centering
    \begin{subfigure}[t]{0.48\linewidth}
        \centering
        \includegraphics[width=\linewidth]{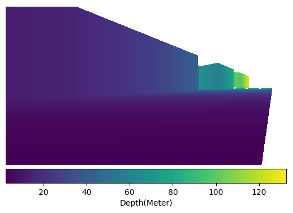}
        \caption{Depth map includes the distances to objects in the scene.}
        \label{fig:raycasting-depth}
    \end{subfigure}
    \hfill
    \begin{subfigure}[t]{0.48\linewidth}
        \centering
        \includegraphics[width=\linewidth]{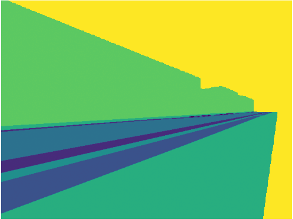}
        \caption{Semantic map represents scene segmentation, where each color encodes a distinct object class.}
        \label{fig:raycasting-semantic}
    \end{subfigure}
    \begin{subfigure}[t]{0.48\linewidth}
        \centering
        \includegraphics[width=\linewidth]{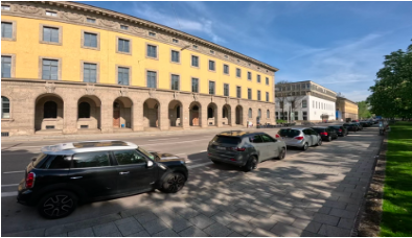}
        \caption{Camera image.}
        \label{fig:raycasting-image}
    \end{subfigure}
    \hfill
    \begin{subfigure}[t]{0.48\linewidth}
        \centering
        \includegraphics[width=\linewidth]{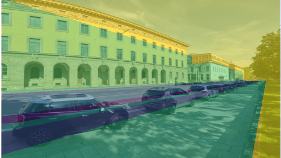}
        \caption{Semantic map overlaid on the image.}
        \label{fig:raycasting-overlay}
    \end{subfigure}
    \caption{Visualizations of the raycasting process, including the generated depth map, semantic map, and camera image. The overlay of the semantic map on the camera image demonstrates alignment and accuracy.}
    \label{fig:raycasting}
\end{figure}

We perform raycasting on the semantic 3D city models using Open3D \cite{zhou2018open3d}. Furthermore, we use a point cloud scan of the \ac{tum} area to determine the precise location and orientation of the device. The generated depth and semantic maps from raycasting are presented in \cref{fig:raycasting-depth} and \cref{fig:raycasting-semantic}. To validate the precision, the semantic map is overlaid on the camera image (\cref{fig:raycasting-overlay}).

\cref{fig:raycasting-depth} and \cref{fig:raycasting-semantic} comprehensively demonstrate the raycasting results. However, in practice, we apply a depth-based mask to exclude semantic and depth information beyond 35 meters, which corresponds to the radar's maximum detection range. This ensures that \ac{sdm} contain only the most relevant information for radar object detection.

\subsection{Fusion Ablation Studies}
Common fusion methods for radar-image data include addition, mean and multiplication, concatenation, and attention mechanism \cite{yao2023radar}. 
These techniques inspire our exploration of radar-\ac{sdm} fusion using addition and concatenation methods.

\begin{figure}[tbh]
    \centering
    \begin{subfigure}{\linewidth}
        \centering
        \begin{subfigure}[t]{0.32\textwidth}
            \centering
            \includegraphics[width=\textwidth]{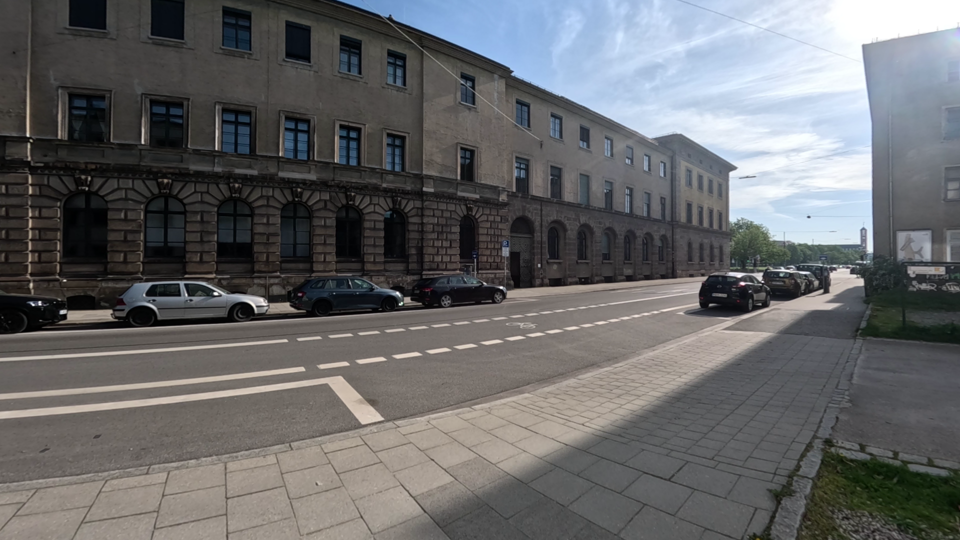}
            \caption{Camera Image}
        \end{subfigure}
        \hfill
        \begin{subfigure}[t]{0.32\textwidth}
            \centering
            \includegraphics[width=\textwidth]{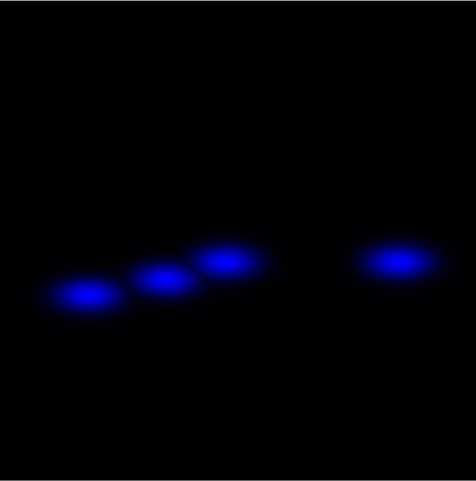}
            \caption{GT \ac{confmaps}}
            \label{fig:feature-fusion-gt}
        \end{subfigure}
        \hfill
        \begin{subfigure}[t]{0.32\textwidth}
            \centering
            \includegraphics[width=\textwidth]{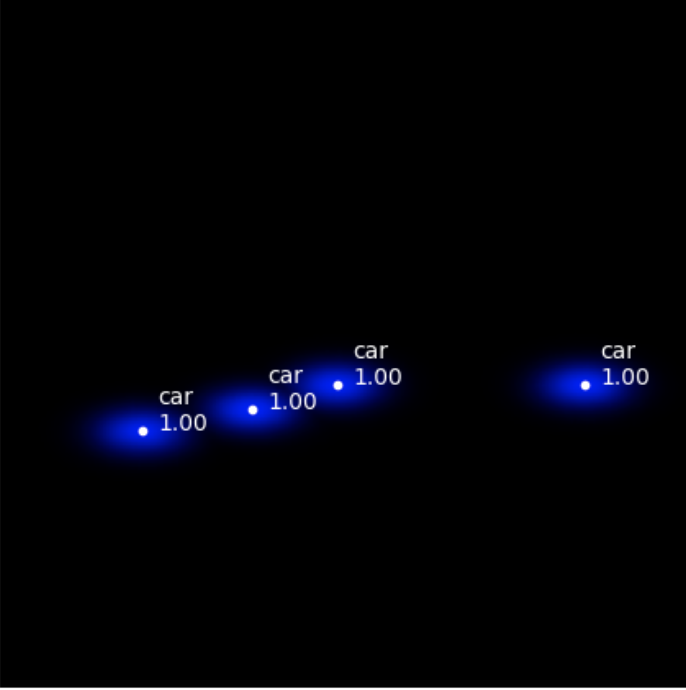}
            \caption{Without \acs{sdm}}
            \label{fig:feature-fusion-without}
        \end{subfigure}
    \end{subfigure}
    \par\bigskip
    \begin{subfigure}{\linewidth}
        \centering
        \begin{subfigure}[t]{0.48\textwidth}
            \centering
            \includegraphics[width=0.958\textwidth]{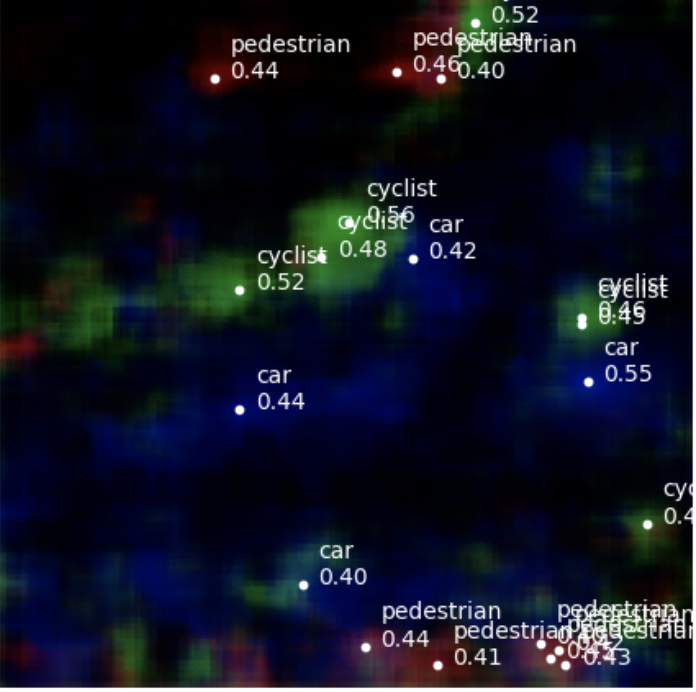}
            \caption{Element-wise Addition}
            \label{fig:feature-fusion-addition}
        \end{subfigure}
        \hfill
        \begin{subfigure}[t]{0.48\textwidth}
            \centering
            \includegraphics[width=0.95\textwidth]{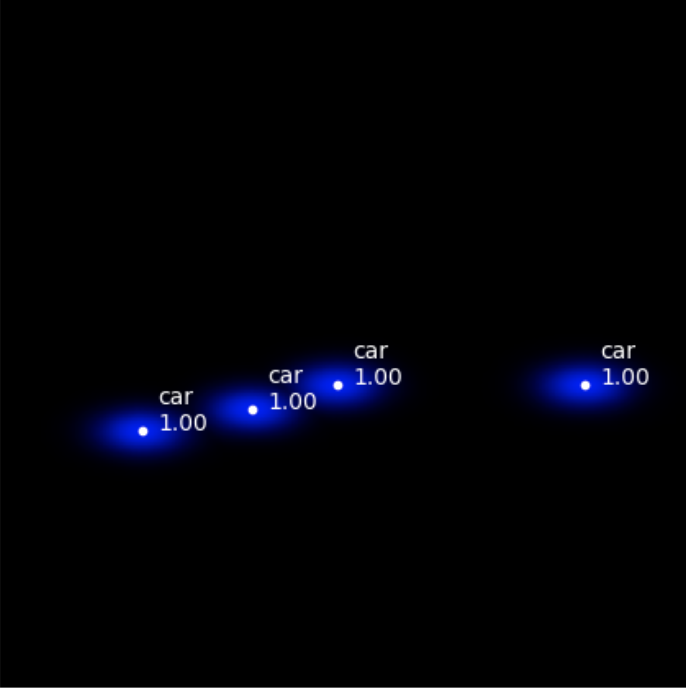}
            \caption{Channel-wise Concatenation}
            \label{fig:feature-fusion-concatenate}
        \end{subfigure}
    \end{subfigure}
    \caption{Visualization of the \acs{confmaps} using different radar-\ac{sdm} fusion methods.}
    \label{fig:feature-fusion}
\end{figure}

To evaluate the effectiveness of different fusion strategies, we visualize the predicted \ac{confmaps} generated by our RADLER on the unseen test data, as shown in \cref{fig:feature-fusion}. 
Detected objects are marked by white dots, each labeled with the predicted object class and confidence score. 
When we perform radar-\ac{sdm} fusion using element-wise addition, unexpected ghost targets appear throughout the predicted \ac{confmaps}.
These artifacts likely arise because adding \ac{sdm} overly influences the radar object detection process. 
This leads to spurious predictions, indicating that element-wise addition may not be suitable for fusing radar and \ac{sdm} features in this context.

We then experiment with channel-wise concatenation. 
By empirically fine-tuning the number of channels to be concatenated in each data modality, we find that this fusion method produces more reliable results. 
\cref{fig:feature-fusion-concatenate} demonstrates that channel-wise concatenation effectively preserves the prediction quality while integrating additional information from \ac{sdm}.

\begin{figure}[tbh]
    \centering
    \includegraphics[width=.9\linewidth]{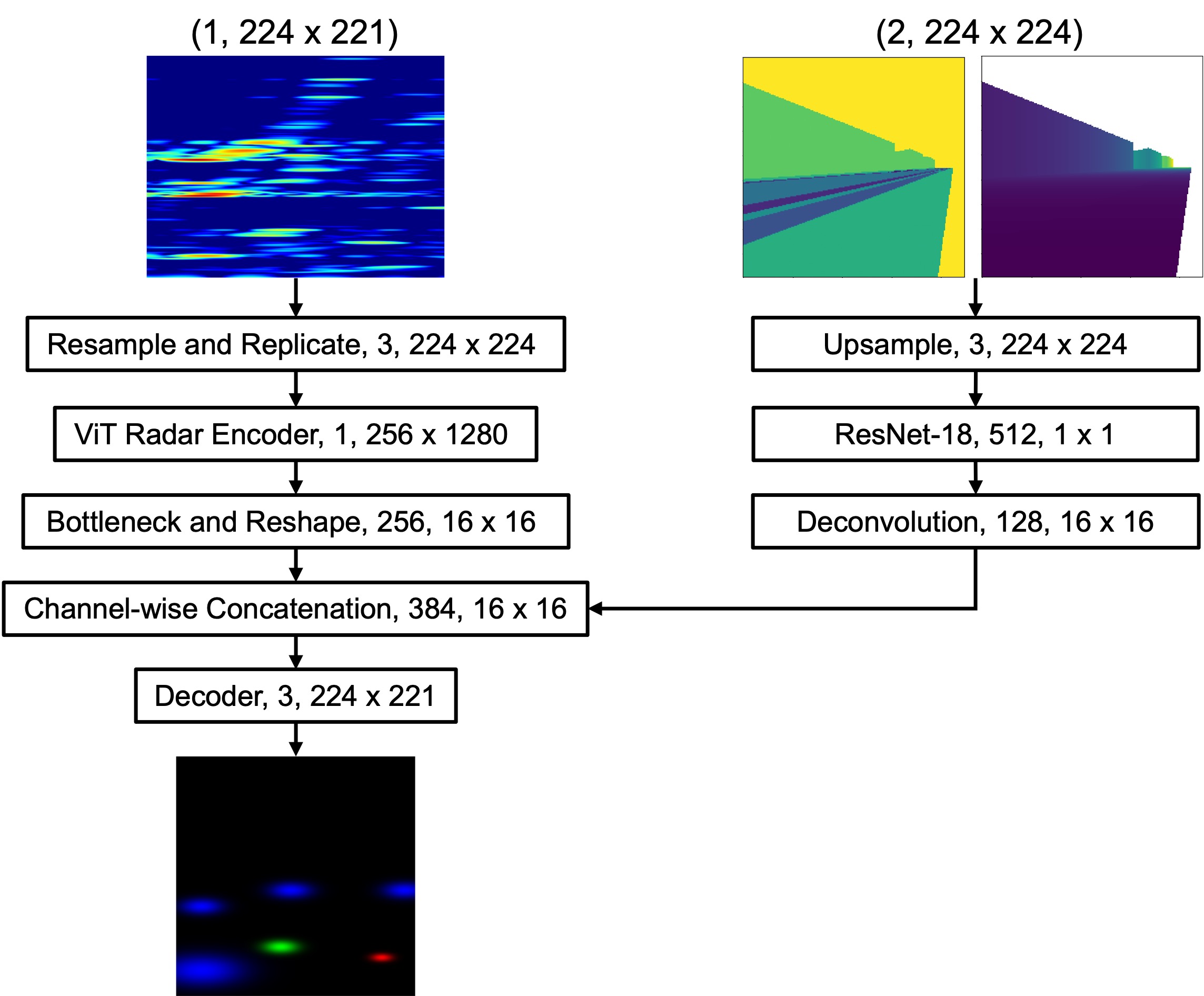}
    \caption{The finalized configuration of RADLER. The data flow indicates transformations in channel size and spatial dimensions at each stage.}
    \label{fig:model-configuration}
\end{figure}

Although channel-wise concatenation appears promising based on visual inspection, a conclusive evaluation of RADLER with the finalized configuration shown in \cref{fig:model-configuration}, using \ac{map} and \ac{mar}, is required to quantify the effectiveness of the radar-\ac{sdm} fusion approach.

\subsection{Radar Object Detection Results}

\begin{table*}[tbh]
    \centering
    \caption{Performance evaluation on the RadarCity dataset. \(\text{\ac{ap}}^\text{\ac{ols}}\) and \(\text{\ac{ar}}^\text{\ac{ols}}\) represent precision and recall under a specific \ac{ols} threshold.}
    \begin{tabular}{c||C{0.8cm}|C{0.8cm}C{0.8cm}C{0.8cm}||C{0.8cm}|C{0.8cm}C{0.8cm}C{0.8cm}}
        \hline
        \rule{0pt}{2.25ex}
        Models & \acs{map} & \(\text{\acs{ap}}^{0.5}\) & \(\text{\acs{ap}}^{0.7}\) & \(\text{\acs{ap}}^{0.9}\) & \acs{mar} & \(\text{\acs{ar}}^{0.5}\) & \(\text{\acs{ar}}^{0.7}\) & \(\text{\acs{ar}}^{0.9}\)\\
        \hline
        RADLER (Without \acs{sdm}) & 90.69 & 91.94 & 90.93 & 88.87 & 95.28 & 95.88 & 95.58 & 93.49\\
        \hline
        RADLER (With \acs{sdm}) & \textbf{94.86} & \textbf{95.89} & \textbf{95.41} & \textbf{92.68} & \textbf{95.95} & \textbf{96.64} & \textbf{96.23} & \textbf{94.19}\\
        \hline\hline
        \acs{rodnet}-CDC & 89.31 & 89.56 & 89.45 & 88.21 & 92.87 & 93.33 & 93.14 & 91.49\\
        \hline
        \acs{rodnet}-HG & 89.74 & 90.46 & 90.21 & 87.72 & 95.26 & 95.96 & 95.51 & 93.63 \\
        \hline
        \acs{rodnet}-HGwI & 83.14 & 83.75 & 83.56 & 81.38 & 89.17 & 89.75 & 89.48 & 87.65\\
        \hline
    \end{tabular}
    \label{tab:evaluation-crctum}
\end{table*}

From \ac{confmaps} to target lists, we adopt the same method used by \ac{rodnet}, namely \ac{lnms}. \ac{lnms} tries to ensure each object is represented by one detection and hence to reduce false positives. It first detects all the peaks, defined as local maxima in a $3\times3$ sliding window, in each channel of the \ac{confmaps}. A peak threshold filters out any peaks with a lower confidence score than the threshold. The \ac{ols} values among the remaining peaks are then calculated, which is defined mathematically as follows:
\begin{equation}\label{math:ols}
    \text{OLS} = \exp \left\{ \frac{-d^2}{2 \left( s \kappa_{\text{cls}} \right)^2} \right\}
\end{equation} where \(d\) is the distance between two detections, \(s\) is the detection's distance to the device, and \(\kappa_{\text{cls}}\) is a per-class constant which represents the error tolerance for object class $\text{cls}$. If the \ac{ols} value of two peaks exceeds the specified threshold, the one with the lower confidence score is suppressed, as they are likely to represent the same object. Otherwise, they are retained as valid detections for different objects. This threshold is empirically chosen to ensure the best performance of our RADLER and \ac{rodnet}.

With target lists, \ac{map} and \ac{mar} are calculated. Another use case of \ac{ols} values is determining whether a prediction is positive or negative. The \ac{ols} values between the predicted results and the \ac{gt} annotations are calculated. Then, different \ac{ols} thresholds are applied, from 0.5 to 0.9, with a step size of 0.05. The \ac{map} and \ac{mar} are the average values of \ac{ap} and \ac{ar} under all \ac{ols} thresholds within this range. We compare RADLER with three \ac{rodnet} models, 3D Convolution-Deconvolution (\ac{rodnet}-CDC), 3D stacked hourglass (\ac{rodnet}-HG), and 3D stacked hourglass with temporal inception (\ac{rodnet}-HGwI). \cref{tab:evaluation-crctum} shows the quantitative evaluation results. \(\text{\ac{ap}}^\text{\ac{ols}}\) and \(\text{\ac{ar}}^\text{\ac{ols}}\) represent precision and recall under a specific \ac{ols} threshold.

Our RADLER, fused with \ac{sdm}, achieves the highest performance among all models, with significant gains in both \ac{map} and \ac{mar}. Compared to \ac{rodnet}-CDC, RADLER achieves relative improvements of 6.22\% in \ac{map} and 3.31\% in \ac{mar}, demonstrating robust detection capabilities across various \ac{ols} values. It also shows improvements of 5.46\% in \ac{map} and 3.51\% over the average performance of \ac{rodnet} models. Even without \ac{sdm}, RADLER outperforms \ac{rodnet}-CDC, highlighting the effectiveness of contrastive \ac{ssl} on radar-image data. Among baseline models, \ac{rodnet}-HG slightly improves upon \ac{rodnet}-CDC in recall but still falls short of RADLER's performance. Conversely, \ac{rodnet}-HGwI shows the weakest results, reinforcing RODNet's limitations in radar object detection and underscoring the advantages of RADLER.

\begin{figure*}[tbh]
    \centering
    \includegraphics[width=\linewidth]{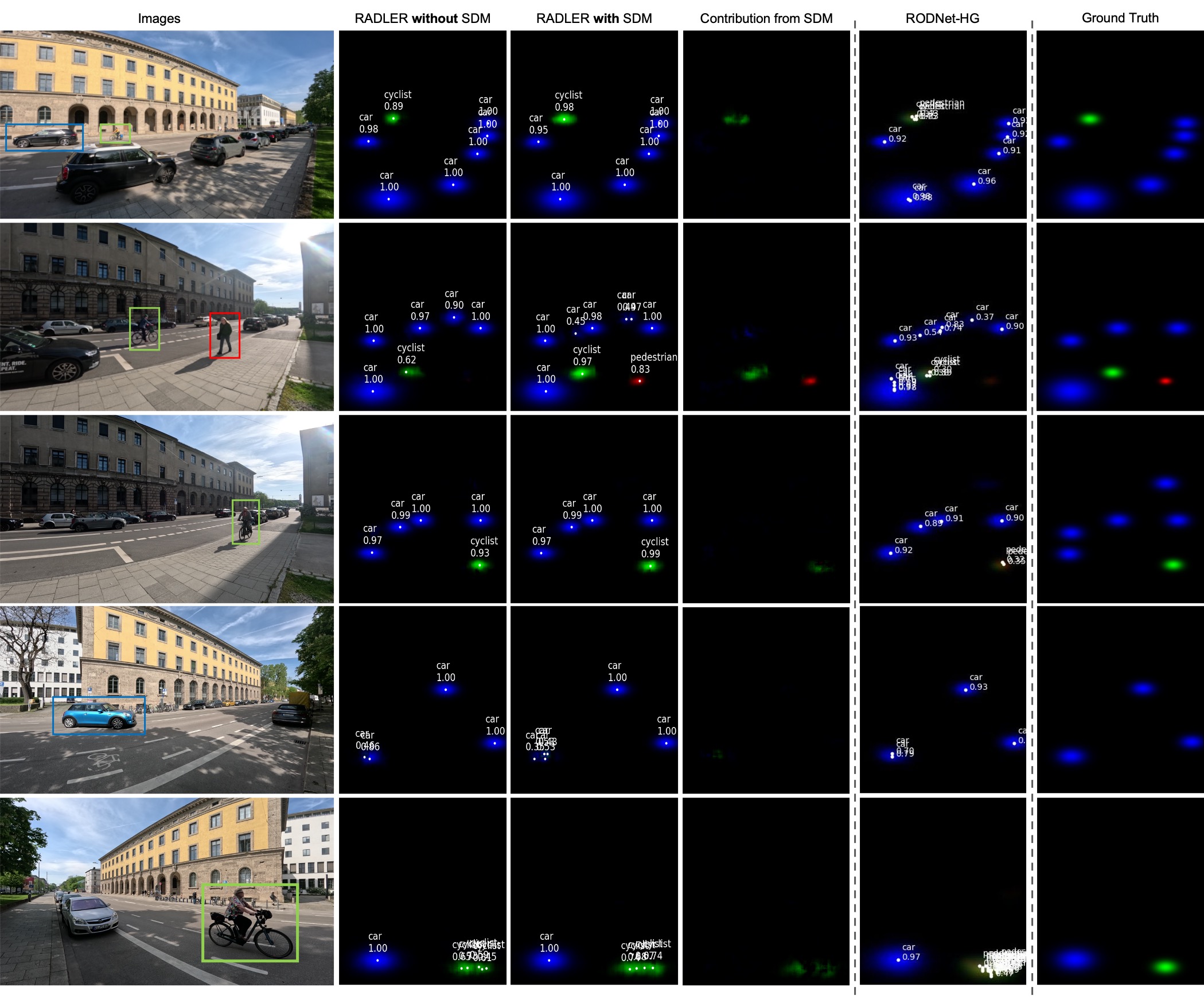}
    \caption{Example detection results from our RADLER and \ac{rodnet}. Object classes (pedestrians, cyclists, and cars) are indicated on the \ac{confmaps} in red, green, and blue, respectively. The \nth{1} column displays the camera images with only selected class instances to improve readability. The \nth{2} column shows detection results from our RADLER without incorporating \acs{sdm}, while the \nth{3} column presents results with \acs{sdm}. The \nth{4} column highlights the impact of \acs{sdm} by visualizing the differences between the \ac{confmaps} of RADLER with and without \acs{sdm}. The \nth{5} column displays detection results from \acs{rodnet}-HG. The last column shows the \ac{gt} \ac{confmaps}.}
    \label{fig:visual-results}
\end{figure*}

Our RADLER's advantage over \ac{rodnet} is further demonstrated in the visualized results (\cref{fig:visual-results}). \ac{rodnet}-HG, as the best-performing \ac{rodnet} variant, tends to produce duplicate detections for the same object. In contrast, our RADLER, both with and without \ac{sdm}, produces fewer duplicate detections with higher confidence scores, suggesting more distinct and localized object representations on the \ac{confmaps}. For example, in the \nth{3} row in \cref{fig:visual-results}, RADLER detects the cyclist with confidence scores exceeding 0.9, while \ac{rodnet}-HG produces detection with a confidence score of only around 0.35. In the last row, RADLER can effectively reduce duplicate detection compared to \ac{rodnet}-HG in cases where the object is close to our device.

\subsection{Ablation Study}
By comparing the first two rows in \cref{tab:evaluation-crctum}, the performance gain by introducing \ac{sdm} in RADLER is significant. Incorporating \ac{sdm} results in a 4.17\% improvement in \ac{map} and a 0.67\% increase in \ac{mar} compared to RADLER without \ac{sdm}. The performance boost is consistent across different \ac{ols} values, indicating a steady contribution from \ac{sdm}. These results confirm the effectiveness of incorporating semantic-depth features into \ac{ra} maps and demonstrate a clear advantage over existing \ac{rodnet} models. 

The claim is further evidenced in \cref{fig:visual-results} where \ac{sdm} has increased the confidence scores of the cyclists in all three examples. In the second example, the pedestrian is detected by RADLER fusing \ac{sdm} with a high confidence score, to which \ac{sdm} contribute substantially. Moreover, \ac{sdm} aids in detecting the car in the middle of three parked vehicles across the street, which RADLER without \ac{sdm} fails to identify. The contribution from \ac{sdm} is also apparent when visualizing the difference between \ac{confmaps} from RADLER with and without \ac{sdm}. 

However, \ac{sdm} do not always contribute to the detection. In the first and second examples, cars detected by RADLER with \ac{sdm} have slightly lower confidence scores. Despite these occasional limitations, the overall advantage of fusing \ac{sdm} is clearly demonstrated.

\subsection{Evaluation on the CRUW dataset}
The autoencoder-based \ac{rodnet}-CDC's encoder and decoder are detachable. We train the encoder of \ac{rodnet}-CDC using the same method shown in \cref{fig:model-architecture}. To validate the viability of contrastive \ac{ssl} on radar-image data, we compare the performance of \ac{rodnet}-CDC trained in both supervised and self-supervised manners on the CRUW dataset.

\begin{table}[tbh]
    \centering
    \caption{Performance evaluation of \ac{rodnet}-CDC trained in different ways on the CRUW dataset.}
    \begin{adjustbox}{width=\linewidth}
        \begin{tabular}{|c|cc|cc|cc|}
            \hline
            \multirow{2}{*}{Training Method} & \multicolumn{2}{c|}{Peak = 0.3} & \multicolumn{2}{c|}{Peak = 0.8} & \multicolumn{2}{c|}{Peak = 0.9} \\
                                   & mAP & mAR                        & mAP & mAR                        & mAP & mAR                        \\ \hline
            Supervised  & 68.62 & 74.91 & 38.26 & 41.23 & 2.41 & 2.07 \\ 
            Self-supervised & 44.04 & 49.96 & 31.23 & 34.28 & 17.95 & 19.11 \\ \hline
        \end{tabular}
    \end{adjustbox}
    \label{tab:evaluation-cruw}
\end{table}

The results are shown in \cref{tab:evaluation-cruw}. When the peak threshold in \ac{lnms} is low (0.3), \ac{rodnet}-CDC trained with supervised learning shows a noticeable performance advantage over the self-supervised version. However, as the threshold increases to 0.8, the performance gap narrows. At a high peak threshold of 0.9, \ac{rodnet}-CDC trained self-supervised demonstrates a significant performance lead, achieving 15.54\% and 17.04\% improvements in \ac{map} and \ac{mar}, respectively, over its supervised counterpart. 

Peak threshold in \ac{lnms} represents requirements for detection quality, with higher values demanding more precise detections. The self-supervised \ac{rodnet}-CDC's superior performance at high thresholds further supports our findings from the RadarCity dataset: contrastive \ac{ssl} enables the model to produce more distinct and localized object representations on the \ac{confmaps} than supervised training. This highlights the potential of self-supervised learning to improve detection quality, especially for challenging cases requiring high confidence.

The \ac{rodnet}-CDC trained self-supervised shows promise; however, its overall performance may improve with larger training datasets, as contrastive \ac{ssl} benefits from extensive data. Additionally, the architecture of the \ac{rodnet}-CDC is relatively simple and may limit its ability to learn comprehensive representations. Prior studies such as SimCLR \cite{chen2020simple} and MoCo \cite{he2020momentum} have demonstrated that more complex architecture can better utilize contrastive loss to generate richer and more generalizable features. Nevertheless, this study validates the effectiveness of contrastive \ac{ssl} for radar-image data and highlights its potential for further exploration.
\section{Conclusion and Limitation}
\label{sec:conclusion}

This paper introduces RADLER, a novel radar object detection method developed under contrastive \ac{ssl}, which leverages prior information from semantic 3D city models for more accurate and noise-robust performance.

Semantic-depth features extracted from \ac{sdm} are fused with \ac{ra} map features, yielding notable performance improvements (5.46\% in \ac{map} and 3.51\% in \ac{mar} on average). 
We observe that the introduced RadarCity dataset, containing 54K synchronized radar-image data pairs, offers unique and, to date, the only possibility to test and develop semantic-map-based radar object detection methods.
Evaluation on the CRUW dataset further validates the potential of contrastive \ac{ssl} for radar object detection. 
While challenges such as improved data synchronization, automated annotation methods, and group object detection remain, this work highlights the effectiveness of combining semantic 3D city models with radar data. 
It provides a solid foundation for future advancements in radar object detection.

{\bf Acknowledgments}
The work was conducted within the framework of the {Leonhard Obermeyer Center} at \ac{tum}.
We are grateful for the diligent work of the TUM2TWIN members and happily support this open data initiative with our data.
{
    \small
    \bibliographystyle{ieeenat_fullname}
    \bibliography{main}
}

\clearpage
\setcounter{page}{1}
\maketitlesupplementary

\section{RadarCity Dataset}
\subsection{Sensor and Device Setup}

The sensor platform for collecting the dataset contains a GoPro HERO11 Black camera and a 77GHz \ac{fmcw} \ac{mmwave} radar from Texas Instrument, model AWR1843Boost. The radar’s \ac{fov} after post-processing is 1-33.7m, ±60° horizontally. The GoPro camera uses the ultra-wide mode to enable a horizontal \ac{fov} of also ±60°. The frame rate of the camera recording is 30 \ac{fps}, and for the radar, it is 15 \ac{fps}.

\begin{figure}[tbh]
    \centering
    \includegraphics[width=.95\linewidth]{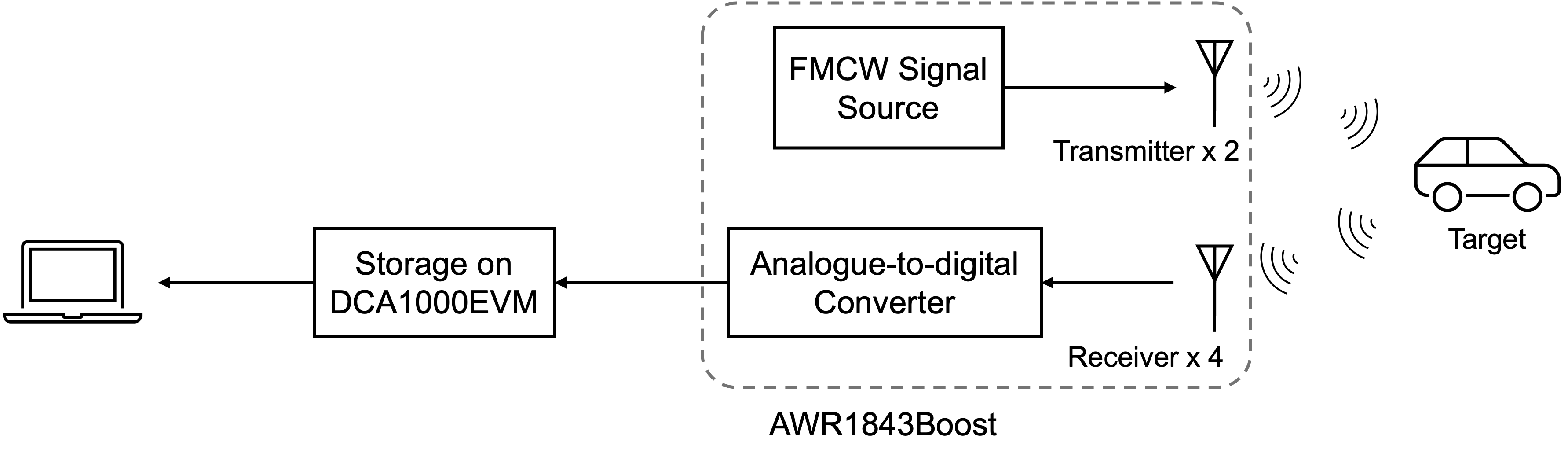}
    \caption{A simplified workflow for the radar data collection. Two transmitters emit signals from the \ac{fmcw} signal source, while four receivers collect the signals reflected by the target. These signals are then converted to digital signals, stored in onboard storage of the DCA1000EVM board, and transmitted to the computer.}
    \label{fig:radar-workflow}
\end{figure}

Attempts have been made to record radar data at 30 \ac{fps} to facilitate data alignment. However, under our radar configuration, the radar data recording could only last around five minutes at 30 \ac{fps} due to limited bandwidth for data transmission. The collected radar data is initially stored in the onboard storage and then transmitted to the computer via Ethernet, as depicted in \cref{fig:radar-workflow}. However, there is a mismatch in the data collection and transmission rate, causing the onboard storage to be filled up after a certain time. Subsequently, any collected data beyond the storage capacity will be abandoned, leading to data loss. To maintain data integrity, it is required to reduce the frame rate of the radar data recording to 15 \ac{fps}.

However, to ensure that every second camera image frame matches a radar frame, it is essential to start recording simultaneously on both the camera and radar. The proposed method uses a synchronized software trigger to start the recording on both devices. For the GoPro camera, a lab firmware is available, offering the functionality to synchronize the built-in clock of the GoPro camera with the computer's system clock. Additionally, the GoPro camera can start recording automatically at a given time. 

The radar board is controlled through mmWave Studio, a GUI from Texas Instruments that allows configuration and control of mmWave sensor modules and collection of analog-to-digital data for offline analysis. A script has been developed to read the computer's system clock so the buttons to trigger the radar data recording can be clicked at a specified time. 

As a result, the GoPro camera and radar can be triggered to record data automatically and simultaneously based on the computer's system clock.

\subsection{Data Clusters}

After collecting the data, they are organized into clusters based on object types and their positions in the scene. \cref{tab:dataclusters} shows detailed information about the data clusters. Data clusters 1 and 2 contain three common objects on the street: pedestrians, cyclists, and cars. These objects are also the targets of interest in many other machine learning models for radar object detection.

\begin{table*}[htpb]
  \caption{Data Clusters in the RadarCity Dataset.}
  \label{tab:dataclusters}
  \centering
  \renewcommand{\arraystretch}{1.5}
  \begin{tabular}{c p{5cm} p{4.5cm} c}
    \toprule
      Data Cluster & Object Classes & Objects' Position & Number of Data Pairs \\
    \midrule
      1 & pedestrians, cyclists, and cars & In the corresponding lanes & \num{38800} \\ \hline
      2 & pedestrians, cyclists, and cars & Not in the corresponding lanes & \num{5300} \\ \hline
      3 & Besides mentioned above, trucks, vans, scooters, motorbikes, and bicycles for delivery, etc. & May or may not be in the corresponding lanes & \num{18700} \\
    \bottomrule
  \end{tabular}
\end{table*}

\begin{figure*}
    \centering
    \begin{subfigure}[t]{.33\linewidth}
        \centering
        \includegraphics[width=\linewidth]{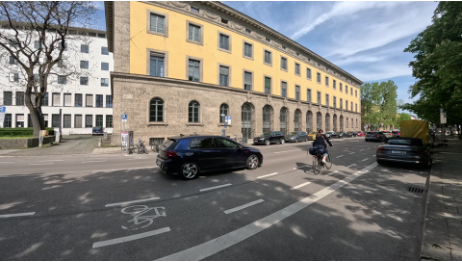}
        \caption{Data Cluster 1: the objects are in the corresponding lanes as the car and cyclist here.}
        \label{fig:data_cluster_1}
    \end{subfigure}
    \hfill
    \begin{subfigure}[t]{.33\linewidth}
        \centering
        \includegraphics[width=\linewidth]{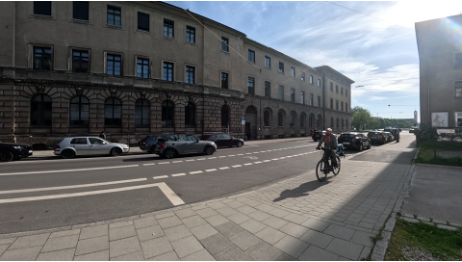}
        \caption{Data Cluster 2: the objects are not in the corresponding lanes.}
        \label{fig:data_cluster_2}
    \end{subfigure}
    \hfill
    \begin{subfigure}[t]{.33\linewidth}
        \centering
        \includegraphics[width=\linewidth]{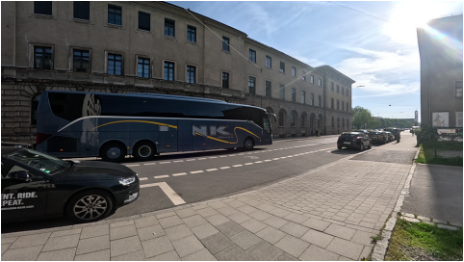}
        \caption{Data Cluster 3: objects other than cars, cyclists, and pedestrians.}
        \label{fig:data_cluster_3}
    \end{subfigure}
    \caption{Examples from different data clusters.}
\end{figure*}

Data cluster 1 and 2 are further split based on where those objects are. \cref{fig:data_cluster_1} shows an example of data cluster 1 where the cars are on the driving lane, and a cyclist is riding on the bicycle lanes. \cref{fig:data_cluster_2}, an example data from data cluster 2, contains the same object classes. However, the cyclist there is riding on the pedestrian walk instead of the bicycle lane. This separation is prepared for further analysis of how to use the prior information from the semantic 3D city models, as the initial thought is that the traffic lanes in the street model should facilitate the detection of common objects traveling on them. Objects, such as the bus in \cref{fig:data_cluster_3}, are out of our current research interest.

\subsection{Dataset Annotation Overview}

\begin{figure}[tbh]
    \centering
    \includegraphics[width=.95\linewidth]{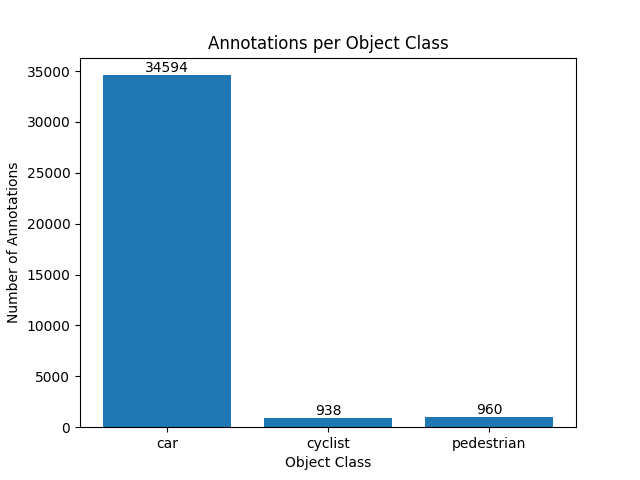}
    \caption{Statistics of the downstream task dataset annotations.}
    \label{fig:annotations}
\end{figure}

Data clusters 1 and 2 were selected for training and evaluating the performance of RADLER against the baseline \ac{rodnet} models. Based on the workflow of contrastive \ac{ssl}, the data in data clusters 1 and 2 are further split into the pretext and downstream task datasets. The pretext task dataset has around \num{35000} radar-image data pairs, while the downstream task dataset has \num{10000} data pairs with \num{8000} for training and \num{2000} for testing. Although \ac{ssl} can save the efforts of annotating the pretext task dataset, the downstream task dataset still needs to be annotated. In this case, the downstream task dataset is purely manually annotated, hence incurring some offsets in the annotations despite great carefulness. The distribution of the annotations of different object classes in the downstream dataset is presented in \cref{fig:annotations}. More cars are annotated than pedestrians and cyclists since those cars are annotated repeatedly among the radar data frames.

\section{Evaluation Results}

\subsection{Impact of OLS in L-NMS}

\begin{figure*}[tbh]
    \centering
    \includegraphics[width=0.95\textwidth]{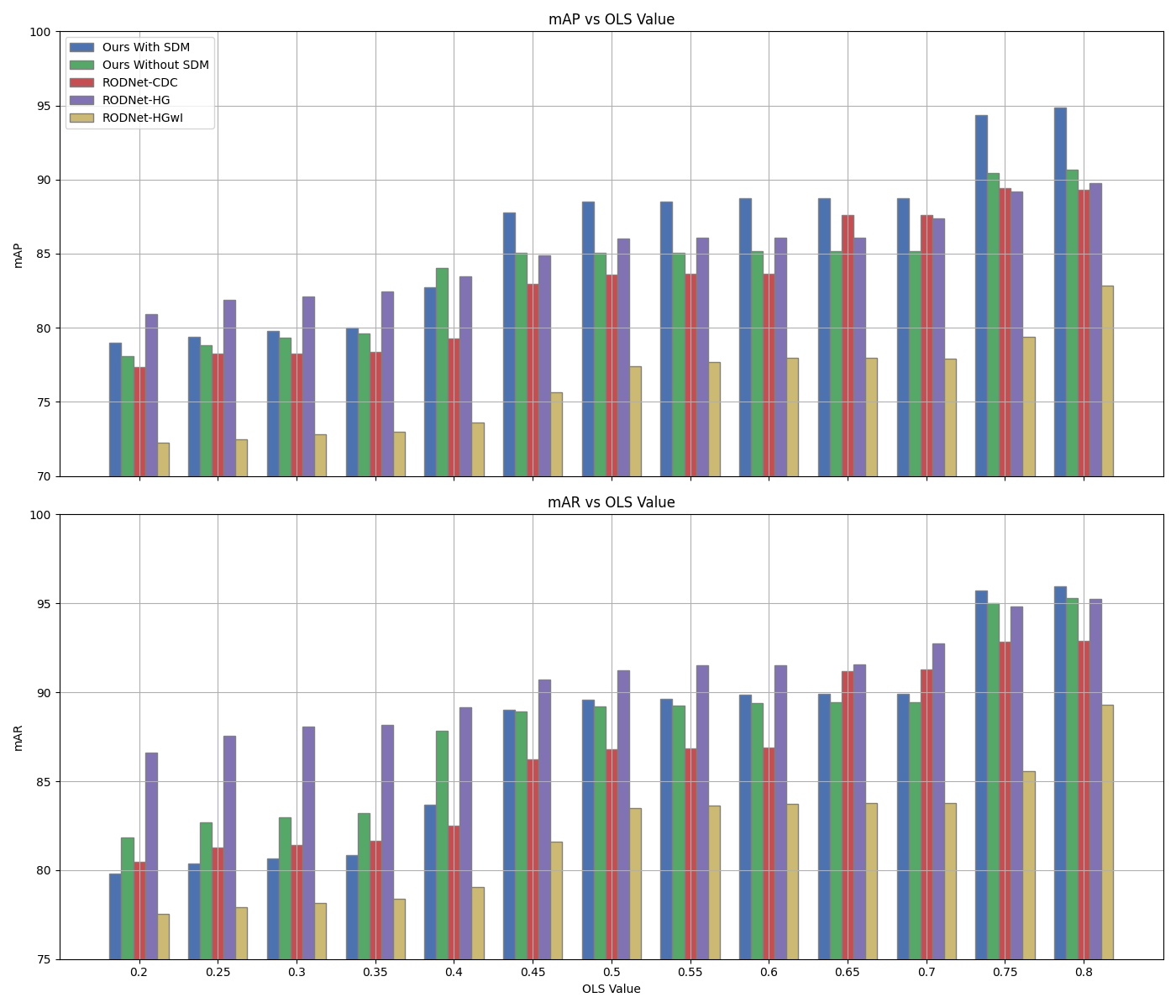}
    \caption{The \ac{map} and \ac{mar} of RADLER and \ac{rodnet} under different \ac{ols} values as the threshold for \ac{lnms}.}
    \label{fig:ap-ar-ols}
\end{figure*}

The \ac{ols} value used in \ac{lnms} is chosen through an experimental evaluation. The results of \ac{map} and \ac{mar} for RADLER and \ac{rodnet} under different \ac{ols} values as thresholds for \ac{lnms} are shown in \cref{fig:ap-ar-ols}. An observation applied for all is that \ac{map} and \ac{mar} improve as the \ac{ols} value increases.

Notably, \ac{rodnet}-HG performs better than other models at lower \ac{ols} values (from \num{0.2} to \num{0.4}). This suggests that \ac{rodnet}-HG produces more separated peaks in the \ac{confmaps}, leading to fewer overlaps and, therefore, lower similarities between the peaks to have them survive the \ac{lnms}. As the \ac{ols} value increases, RADLER, particularly the variant integrated with \ac{sdm}, starts to outperform the others, with its advantage becoming most apparent at an \ac{ols} value of \num{0.8}, where it achieves a \ac{map} of nearly \num{95}\% and a \ac{mar} of around \num{96}\%.

While the exact reasons for this performance trend across different \ac{ols} values may not be immediately clear, this experiment highlights how the distribution of confidence values in the \ac{confmaps} varies across models. The same \ac{ols} threshold applied to different models yields different target lists, indicating the differences in how the confidence values of a detected object from different models are numerically distributed on the \ac{confmaps}, resulting in the quality of the target list generated through \ac{lnms}.

\subsection{Additional Qualitative Results}

\begin{figure*}[tbh]
    \centering
    \includegraphics[width=.95\linewidth]{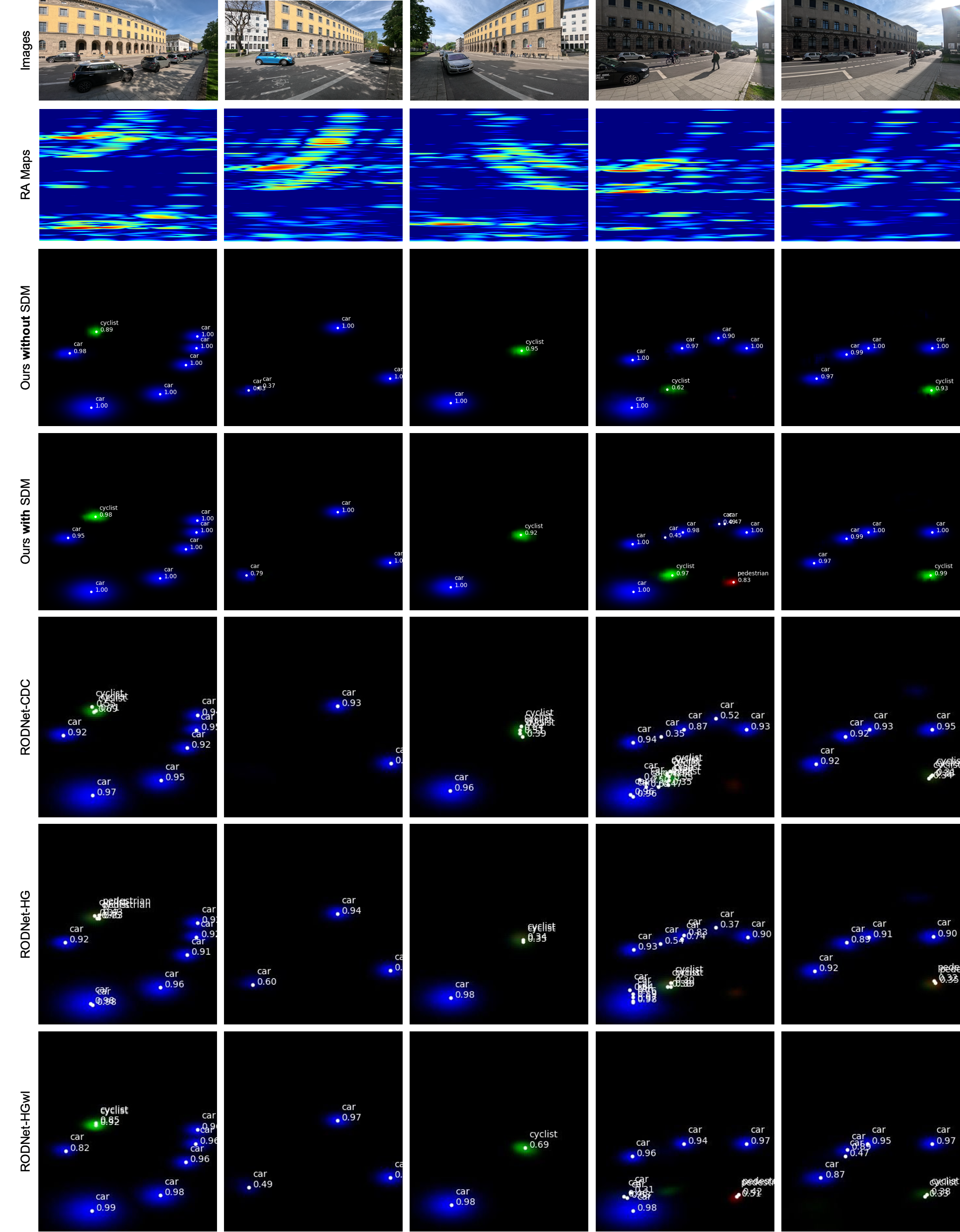}
    \caption{Example results: the first row displays the images, and the second row shows the corresponding \ac{ra} maps. Subsequent rows display predicted \ac{confmaps} from different models.}
    \label{fig:test-results}
\end{figure*}

\cref{fig:test-results} shows more visualized detection results from the test data. Here, the \ac{ols} value for \ac{lnms} is set to be \num{0.8} for all models. RADLER demonstrates a clear advantage over the \ac{rodnet} models. As shown in every column, RADLER can produce a more accurate and confident detection for each object as the \ac{rodnet} models tend to produce redundant detection for one object with lower confidence values.

In the \nth{1} column, \ac{sdm} contribute to the detection of the cyclist by increasing its confidence score by \num{0.09}. Also, in the last column, the usage of \ac{sdm} increases the confidence of the detected cyclist by \num{0.06}.

However, \ac{sdm} are not always contributive. In the \nth{1} column, the detected car on the left in the middle of the street is \num{0.03} lower in confidence score compared to the detection from RADLER without \ac{sdm}. Also, in the \nth{3} column, \ac{sdm} have lowered the confidence of the detected cyclist by \num{0.03}.

\end{document}